\documentclass[journal,transmag,10pt]{IEEEtran}

\usepackage[T1]{fontenc}
\usepackage{graphicx}
\usepackage{breakcites}
\usepackage{url}

\usepackage{booktabs}   %% For formal tables:
\usepackage{subcaption} %% For complex figures with subfigures/subcaptions
\usepackage{listings}
\usepackage{relsize}
\usepackage{enumerate}
\usepackage{float}
\usepackage{nicefrac}
\usepackage{multirow}
\usepackage{microtype}
\usepackage{wrapfig}
\usepackage{amsmath}
\usepackage{amssymb}
\usepackage{mathtools}
\usepackage{xcolor}
\usepackage{amsthm}
\usepackage{wrapfig}
\usepackage{algorithmic}
\usepackage[framemethod=tikz]{mdframed}
\usepackage{gensymb}
\usepackage[tableposition=below]{caption}
\usepackage{mathpartir}
\usepackage[algosection,ruled,vlined,linesnumbered,algo2e]{algorithm2e}
\SetKwInput{KwIn}{Inputs}
  \SetKwFunction{Fbuildassume}{BuildAssumption}
  \SetKwFunction{FSynthSpec}{SynthesizeSpec}
  \SetKwFunction{Ffor}{for}
  \SetKwProg{Fn}{}{:}{}
  \DontPrintSemicolon

\makeatletter
\newcommand{\removelatexerror}{\let\@latex@error\@gobble}
\makeatother

\newtheorem{theorem}{Theorem}

\newtheorem{definition}[theorem]{Definition}

\newtheorem{proposition}[theorem]{Proposition}

\definecolor{blue2}{rgb}{0.0, 0.5, 1.0}

\newcommand{\he}{{\tt HE}}
\newcommand{\vcte}{{\tt cte}}
\newcommand{\vhe}{{\tt he}}
\newcommand{\vctec}{\underline{{\tt cte}}}
\newcommand{\vhec}{\underline{{\tt he}}}

\newcommand{\prism}{\textsc{PRISM}}

\definecolor{prismgreen}{rgb}{0, 0.6, 0}

\lstdefinelanguage{Prism}{
    backgroundcolor=\color{white}, % choose your background color
    basicstyle=\scriptsize\ttfamily, % choose your font style and size
    keywords={bool,C,ceil,const,ctmc,double,dtmc,endinit,endmodule,endrewards,endsystem,F,false,floor,formula,G,global,I,init,int,label,max,mdp,min,module,nondeterministic,P,Pmin,Pmax,prob,probabilistic,R,rate,rewards,Rmin,Rmax,S,stochastic,system,true,U,X},
    keywordstyle=\color{blue}\bfseries, % choose your keyword color and style
    comment=[l]{//},
    morecomment=[s]{/*}{*/},
    commentstyle=\color{prismgreen}\itshape, % choose your comment color and style
    stringstyle=\color{purple}, % choose your string color
    showstringspaces=false, % don't show spaces in strings
    breaklines=true, % allow line breaks
    frame=single, % add a frame around the code
    framexleftmargin=1pt, % adjust the frame margin
    captionpos=b % position the caption at the bottom
    escapechar=@, % write LaTeX comments escaped by @ symbol
    tabsize=4,
    literate={->}{$\rightarrow{}$}{1}  {<=}{$\leq$}{1} {>=}{$\geq$}{1} {!=}{$\neq$}{1}  {epsilon}{$\varepsilon$}{1} {M2}{{$M_2$}}{1} {||}{$\parop$}{1}
}

\lstdefinelanguage{LTSA}{
    backgroundcolor=\color{white}, % choose your background color
    basicstyle=\scriptsize\ttfamily, % choose your font style and size
    keywords={const, if, else, then,range},
    keywordstyle=\color{blue}\bfseries, % choose your keyword color and style
    comment=[l]{//},
    morecomment=[s]{/*}{*/},
    commentstyle=\color{gray}\itshape, % choose your comment color and style
    stringstyle=\color{purple}, % choose your string color
    showstringspaces=false, % don't show spaces in strings
    breaklines=true, % allow line breaks
    frame=single, % add a frame around the code
    framexleftmargin=1pt, % adjust the frame margin
    captionpos=b % position the caption at the bottom
    escapechar=@, % write LaTeX comments escaped by @ symbol
    tabsize=4,
    literate={->}{$\rightarrow{}$}{1}  {<=}{$\leq$}{1} {>=}{$\geq$}{1} {!=}{$\neq$}{1}  {epsilon}{$\varepsilon$}{1}
}

\newcommand{\parop}{\mathbin{||}}
\newcommand{\projop}[1]{{\downarrow_{#1}}}
\newcommand{\weakest}[1]{{A_{w}^{#1}}}
\newcommand{\fspsyn}[1]{${\tt {#1}}$}
\newcommand{\interalpha}{\Sigma}

\newcommand{\Determinization}{\texttt{Determinization}}
\newcommand{\CompletionWithSink}{\texttt{CompletionWithSink}}
\newcommand{\ErrorRemoval}{\texttt{ErrorRemoval}}
\newcommand{\BackwardErrorPropagation}{\texttt{BackwardErrorPropagation}}

\begin{document}
%
%\title{Compositional Verification of Autonomous Systems with Machine Learning Components}
%\title{Assumption Generation for Autonomous Systems with Machine Learning Components}
\title{Assumption Generation for the Verification of  Learning-Enabled Autonomous Systems}
\author{\IEEEauthorblockN{Corina~P\u{a}s\u{a}reanu\IEEEauthorrefmark{1,2},
Ravi~Mangal\IEEEauthorrefmark{2},
Divya~Gopinath\IEEEauthorrefmark{1},
and
Huafeng Yu\IEEEauthorrefmark{3}}
\IEEEauthorblockA{\IEEEauthorrefmark{1}KBR, NASA Ames~~
\IEEEauthorrefmark{2}Carnegie Mellon University~~
\IEEEauthorrefmark{3}Boeing Research and Technology}}

\maketitle              % typeset the header of the contribution

\begin{abstract}

Providing safety guarantees for autonomous systems is difficult as these systems operate in complex environments that require the use of learning-enabled components, such as deep neural networks (DNNs) for visual perception. DNNs are hard to analyze due to their size (they can have thousands or millions of parameters), lack of formal specifications (DNNs are typically learnt from labeled data, in the absence of any formal requirements), and sensitivity to small changes in the environment. We present an assume-guarantee style compositional approach for the formal verification of system-level safety properties of such autonomous systems. Our insight is that we can analyze the system {\em in the absence} of the DNN perception components by automatically synthesizing {\em assumptions} on the DNN behaviour that {\em guarantee} the satisfaction of the required safety properties. The synthesized assumptions are the {\em weakest} in the sense that they characterize the output sequences of all the possible DNNs that, plugged into the autonomous system, guarantee the required safety properties.  The assumptions can be leveraged as run-time monitors over a deployed DNN to guarantee the safety of the overall system; they can also be mined to extract local specifications for use during training and testing of DNNs. We illustrate our approach on a case study taken from the autonomous airplanes domain that uses a complex DNN for perception.
\end{abstract}
%
%
%
%%
%% Keywords. The author(s) should pick words that accurately describe
%% the work being presented. Separate the keywords with commas.

\section{Introduction}
\label{sec:intro}
%% from old paper: to re-write

Autonomy is increasingly prevalent in many applications, such as recommendation systems, social robots and self-driving vehicles, that require strong safety guarantees. However, this is difficult to achieve, since autonomous systems are meant to operate in uncertain environments that require using machine-learnt components. For instance, deep neural networks (DNNs) can be used in autonomous vehicles to perform complex tasks such as perception from high-dimensional images. DNNs are massive (with thousands, millions or even billions  of parameters) and are inherently opaque, as they are trained based on data, typically in the absence of any specifications, thus precluding formal reasoning over their behaviour.
 %Current assurance approaches involve design and testing procedures that are expensive and inad- equate, as they have been developed mostly for human-in-the-loop systems and do not apply to systems with advanced autonomy.
Current system-level assurance techniques that are based on formal methods, either do not scale to systems that contain complex DNNs~\cite{santa2022nnlander,P.Habeeb2023}, provide no guarantees~\cite{katz2022verification}, or provide only probabilistic guarantees~\cite{pasareanu2023closedloop,hsieh2022verifying} for correct operation of the autonomous system. Falsification techniques \cite{ghosh2021counterexample} can be used to find counterexamples to safety properties but they cannot guarantee that the properties hold.

%\corina{added this paragraph; needs polishing?} 
Moreover, it is known that, even for well-trained, highly-accurate DNNs, their performance degrades in the presence of distribution shifts or adversarial and natural perturbations from the environment (e.g., small changes to correctly classified inputs that cause DNNs to mis-classify them)~\cite{huang2020survey}. 
These phenomena present safety concerns but it is currently unknown how to provide strong assurance guarantees about such behaviours. Despite significant effort in the area, current formal verification and certification techniques for DNNs~\cite{KatzHIJLLSTWZDK19,abs-1710-00486} only scale to modest-sized networks and provide only partial guarantees about input-output DNN behaviour, i.e. they do not cover the whole input space. Furthermore, it is unknown how to relate these (partial) DNN guarantees to strong guarantees about the safety of the overall autonomous system. 

We propose a compositional verification approach for learning-enabled autonomous systems to achieve {\em strong assurance guarantees}. The inputs to the approach are: the design models of an autonomous system, which contains both conventional components (e.g., controller and plant modeled as labeled transition systems) and the learning-enabled components (e.g., DNN used for perception), and a safety property specifying the desired behaviour of the system. 

While the conventional components can be modeled and analyzed using well-established techniques (e.g., using model checking for labeled transition systems, as we do in this paper), the challenge is to reason about the perception components. This includes the complex DNN together with the sensors (e.g., cameras) that generate the high-dimensional DNN inputs (e.g., images), 
which are subject to random perturbations from the environment (e.g., change in light conditions),  %These sensors are complex, with their outputs subject to system state as well as random background conditions, 
all of them difficult, if not impossible, to model precisely.
%together with the sensors and the random perturbations from the environment, which are difficult, if not impossible, to model precisely. 
%\ravi{I think this point maybe difficult to grasp. First, I am not sure if a reader will appreciate why we need models of the sensors and the environment. Second, we do have a precise model of the DNN but not of the sensors or the environment. }
%
%Furthermore, the sheer size of modern DNN makes formal verification intractable. 
To address this challenge, we 
take an abductive reasoning approach, where we analyze the system {\em in the absence} of the DNN and the sensors, deriving conditions on DNN behaviour that guarantee the safety of the overall system. We build on our previous work on automated assume-guarantee  compositional verification~\cite{PasareanuGBCB08, ASE02},  to automatically generate  {\em assumptions} in the form of labeled transition systems, encoding sequences of DNN predictions that guarantee system-level safety. The assumptions are the weakest in the sense that they characterize the output sequences of all the possible DNNs that, plugged into the autonomous system, satisfy the property. We further propose to mine the assumptions  to extract local properties on DNN behavior, which in turn can be used for the separate testing and training of the DNNs.

We envision the approach to be applied at different development phases for the autonomous system. At design time, the approach can be used to uncover problems in the autonomous system {\em before} deployment. 
The automatically generated assumptions and the extracted local properties can be seen as {\em safety requirements} for the development of neural networks.  
At run time, the assumptions can be deployed as safety monitors over the DNN outputs to {\em guarantee the safety behaviour} of the overall system.

%Developers can use the assumptions to test and debug their DNNs, e.g., on sequences of images obtained from simulations. The assumptions can also be used for training of the DNNs on inputs that do not require manual labelling. The extracted local specifications can also be used to facilitate training and testing of DNNs, even for data that does not come in sequence.

%\divya{Highlight the novelty and specific contributions of this work over \cite{ASE02}}.

We summarize our contributions as follows:
%\begin{enumerate}
%\item
(1) Analysis with strong safety guarantees for autonomous systems with learning-enabled perception components. The outcome of the analysis is in the form of {\em assumptions} and {\em local specifications} over DNN behavior, which can be used for training and testing the DNN and also for run-time monitoring to provide the safety guarantees.
%, leveraging the assumption-generation method from~\cite{ASE02}. The outcome of the analysis is in the form of {\em assumptions} over the DNN ``predictions'' that guarantee safety of the system. The assumptions are ``the most permissive'', i.e. by construction they allow all the possible prediction sequences for the DNN that ensure system safety. We also show how the assumptions can be leveraged to extract local properties on DNN behavior, which in turn can be used for the separate testing and training of the DNN. Furthermore, the assumptions can be used as run-time monitors to guarantee the safety of the system during operation.
%\item
(2) Demonstration of the approach on a case study inspired by a realistic scenario of an autonomous taxiing system for airplanes, that uses a complex neural network for perception.
%\item 
(3) Experimental results showing that the extracted assumptions are small and understandable, even if the perception DNN has large output spaces, making them amenable for training and testing of DNNs and also for run-time monitoring. 
%\item 
(4)~Probabilistic analysis, using empirical probabilities derived from profiling the perception DNN, to measure the probability that the extracted assumptions are violated when deployed as run-time safety monitors. Such an analysis enables developers to estimate how restrictive the safety monitors are in practice. 

\section{Preliminaries}
\label{sec:background}
%\subsection
\paragraph{\textbf{Labeled Transition Systems}}
%%from some other paper need to update
We use finite labelled transition systems (LTSs) to model the behaviour of an autonomous system.
%as communicating processes.
%We use LTSA~\cite{ltsa} for analysis.
%Let $\Act$ be the universal set of \emph{observable actions} and let $\tau$ denote a local, unobservable action. 
A \emph{labeled transition system} (LTS) is a
tuple $M = (Q,\Sigma, \delta, q_0)$, where
\begin{itemize}
\item $Q$ is a finite set of \emph{states};
\item $\Sigma$, the \emph{alphabet} of $M$, is a set of
  observable actions;
\item $\delta \subseteq Q \times (\Sigma \cup \{\tau\})  \times Q$ is a \emph{transition
    relation};
\item $q_0 \in Q$ is the initial state.
\end{itemize}
Here $\tau$ denotes a local, unobservable action. We use $\alpha(M)$ to denote the alphabet of $M$ (i.e. $\alpha(M) =\Sigma$).
 %\ravi{Why do we need $\alpha M$ notation? Cant we just use $\Sigma$?}\corina{this is useful to refer to the alphabet of an lts}
 A trace $\sigma\in \Sigma^*$ of an LTS $M$ is a sequence of observable actions that $M$ can perform starting in the initial state. The \emph{language} of $M$, denoted $L(M)$, is the set of traces of $M$. Note that our definition allows non-deterministic transitions.

Given two LTSs $M_1$ and $M_2$, their \emph{parallel composition} $M_1 \parop M_2$ synchronizes shared actions and interleaves the remaining actions. We provide the definition of $\parop$ (which is commutative and associative) in a process-algebra style. Let $M = (Q,\Sigma, \delta, q_0)$ and $M' = (Q',\Sigma', \delta', q'_0)$ be two LTSs. We say that $M$ {\em transits} to $M'$ with action $a$, written as $M \xrightarrow{a}{} M'$, iff $(q_0,a,q_0')\in \delta$, $\Sigma=\Sigma'$, and $\delta=\delta'$. 
Let $M_1 = (Q_1,\Sigma_1, \delta_1, q_{1,0})$ and $M_2 = (Q_2,\Sigma_2, \delta_2, q_{2,0})$. $M_1\parop M_2$ is an LTS $M=(Q,\Sigma, \delta, q_0)$
such that $Q=Q_1 \times Q_2$, $q_0=(q_{1,0},q_{2,0})$, $\Sigma = \Sigma_1 \cup \Sigma_2$ and $\delta$ is defined as follows:
\begin{mathpar}
  \inferrule{M_1 \xrightarrow{a}{} M_1', a\not\in\Sigma_2}{M_1 \parop M_2 \xrightarrow{a}{} M_1'\parop M_2} \and
  \inferrule{M_1 \xrightarrow{a}{} M_1', M_2 \xrightarrow{a}{} M_2', a\neq\tau}{M_1 \parop M_2 \xrightarrow{a}{} M_1'\parop M_2'}
\end{mathpar}

%\begin{mathpar}
%  \inferrule{(q'_1,\sigma,q'_2)\in \delta'~~~
%  (q''_1,\sigma,q''_2)\in \delta''~~~ \sigma\not=\tau}{((q'_1,q''_1), \sigma, (q'_2,q''_2))\in \delta} \and
  %
%  \inferrule{(q'_1,\sigma,q'_2)\in \delta'~~~ 
%  (\sigma\not \in \Sigma'' \vee \sigma=\tau) ~~~ q'' \in Q''}{((q'_1,q''), \sigma, (q'_2,q''))\in \delta} \and
%  \inferrule{(q''_1,\sigma,q''_2)\in \delta''~~~
%  (\sigma\not \in \Sigma' \vee \sigma=\tau)~~~ q' \in Q' }{((q',q''_1), \sigma, (q',q''_2))\in \delta}
%\end{mathpar}

% is an LTS that satisfies
% $L(M_1 \parop M_2) = L(M_1) \cap L(M_2)$.
We also use LTSs to represent safety
properties $P$.  $P$ can be synthesized, for example, from a specification in a temporal logic formalism such as (fluent) LTL \cite{fluents}.
%A safety property is a deterministic LTS \footnote{An LTS is deterministic if $\delta$ is a function instead of a relation and there are no $\tau$ transitions.} .
The language of $P$ describes the set of allowable behaviours for $M$; $M\models P$ iff $L(M\projop {\alpha(P)})\subseteq L(P)$ where $\alpha(P)$ is the alphabet of of $P$.
%\ravi{This is minor but I prefer $\alpha_P$ over $\alpha P$}. 
The  $\projop \Sigma$ operation hides (i.e., makes unobservable by replacing with $\tau$) all the observable actions of an LTS that are not in $\Sigma$. 
%\ravi{Perhaps should use a different letter than $A$ since we use $A$ later for assumption.} 
The verification of property $P$ is performed by first building an {\em error} LTS, $P_{\mathit{err}}$, which is the complement of $P$  trapping possible violations with an extra error state $\mathit{err}$, and checking reachability of $\mathit{err}$ in $M\parop P_{\mathit{err}}$.

%\subsection{Assume-Guarantee Reasoning}
%Assume-guarantee reasoning decomposes the verification of a large system into the verification of its components, using {\em assumptions} about their context. The simplest rule for such reasoning checks if a system composed of $M_1$ and $M_2$ satisfies a property $P$ by checking that $M_1$ satisfies $P$ under assumption $A$ and discharging $A$ on 'context' $M_2$. This rule can be represented as follows:

%\begin{mathpar}
%  \inferrule{\langle A,M_1,P \rangle \wedge 
%  \langle true,M_2,A \rangle}{\langle true,M_1\|M_2,P \rangle} 
%\end{mathpar}

%Here $\langle A,M,P \rangle$ denotes an assume-guarantee triple which is true if whenever $M$ is part of a system that satisfies $A$ that system also satisfies $P$.

%\subsection
\paragraph{\textbf{Weakest Assumption}}
\label{sec:background_assumption}
%We leverage the algorithm from previous work~\cite{ASE02} to automatically build the {\em weakest assumption} for a system modeled as an LTS which we formalize here.

%For a system $M$ (modeled as an LTS) to satisfy a property $P$, $M$ is viewed as an {\em open} system interacting with its environment through an interface $\Sigma_I\subseteq \alpha M$. The weakest assumption characterizes {\em all} the {\em environments} in which $M$ is expected to satisfy the property. We formalize it here, generalizing from~\cite{ASE02}.

%\ravi{Reworded a little bit below. Check}
A component (or subsystem) $M$, modeled as an LTS, can be viewed as {\em open}, interacting with its {\em context} (i.e., other components or the external world) through an interface $\Sigma_I\subseteq \alpha(M)$. For a property $P$, the weakest assumption
%~\cite{ASE02} over alphabet $\Sigma_I$ 
characterizes {\em all} the {\em contexts} in which $M$ can be guaranteed to satisfy the property. We formalize it here, generalizing from~\cite{ASE02} to consider a subset of $\Sigma_I$.

\begin{definition}[\textbf{Weakest assumption}] 
For LTS $M$, safety property $P$ ($\alpha(P)\subseteq\alpha(M)$) and $\interalpha\subseteq\Sigma_I$, the weakest assumption $\weakest{\interalpha}$ for $M$ with respect to $P$ and $\interalpha$ is a (deterministic) LTS such that $\alpha(\weakest{\interalpha}) =\interalpha$ and for any other component $N$, $M\projop{\interalpha}\parop N \models P$ iff $N\models \weakest{\interalpha}$.
\end{definition}

\label{sec:assume_gen}
Prior work~\cite{ASE02} describes an algorithm for building the weakest assumption for components and safety properties modeled as LTSs. We modify it for our purpose in this paper.

\section{Compositional Verification of Learning-enabled Autonomous Systems}
\label{sec:verif}
%We illustrate our compositional approach via an example of an autonomous system, namely, an autonomous system for guiding airplanes on taxiways, as 

%\subsection{Overview}
\label{sec:verif_overview}

We present a compositional approach for verifying the safety of autonomous systems with learning-enabled components.  We model our system as a parallel composition of LTSs; however our approach is more general, and can be adapted to reasoning about more complex, possibly unbounded, representations, such as %\ravi{added} 
transition systems with countably infinite number of states and hybrid systems, by leveraging previous work on assuming-guarantee reasoning for such systems~\cite{GiannakopoulouP13,BogomolovFGGPPS14}. 
%Our approach is generally applicable to any autonomous system that can be modeled as an LTS.  \todo{Mention that approach is even more generic; hybrid systems etc.} However, in the rest of this paper, 
We focus on cyber-physical systems that use DNNs for vision perception (or more generally, perception from high-dimensional data). These DNNs are particularly difficult to reason about, due to large sizes, opaque nature, and sensitivity to input perturbations or distribution shifts. 

Let us consider an autonomous system consisting of four components; systems with more components can be treated similarly. The system contains a $Perception$ component (i.e., a DNN) which processes images ($img \in Img$) and produces estimates ($s_{est} \in Est$) of the system state, a $Controller$ that sends commands\footnote{We use ``commands'' instead of ``actions'' since we already use actions to refer to the transition labels of LTSs.} ($c \in Cmd$) to the physical system  being controlled in order to maneuver it based on these state estimates, the $Dynamics$ modeling the evolution of the actual physical system states ($s \in Act$) in response to control signals, and the $Sensor$, e.g., a high-definition camera, that captures images representing the current state of the system and its surrounding environment ($e \in Env$). 
%background ($b \in Bkg$)
%These images depend not only on the system state but are also subject to the state of the background
%conditions of the environment 
%in which the autonomous system operates.
%\ravi{I think $Sensor$ does not give a complete description of whats going on. There is a physical process outside the sensor that generates the photons being captured by the camera based on the system state and the environment, and then a process inside the sensor that translates the analog signal into a digital image.}
There may be other sensors (radar, LIDAR, GPS) that we abstract away for simplicity.

Suppose that each of these components  can be modeled as an LTS. The alphabet of observable actions for each component is as follows: $\alpha(Perception) = Img \cup Est$, $\alpha(Controller) = Est \cup Cmd$, $\alpha(Dynamics) = Act \cup Cmd$, and $\alpha(Sensor) = Act \cup Env \cup Img$.
We can  write the overall system as $System=Sensor\parop Perception \parop Controller \parop Dynamics$.

% Table~\ref{tab:lts_components} describes the type of each of these components when viewed as a function as well as the alphabet of the corresponding LTS. The set of states $Q$ for each corresponding LTS is a subset of natural numbers ($\mathbb{N}$). The key thing to note is that in the LTS encoding the domain and the co-domain of the function become the alphabets.

% \begin{table}[h]
% \centering
% \label{tab:components}
% \begin{tabular}{|c|c|c|}
% \hline
% Component & Function Type & LTS alphabet ($\Sigma$) \\
% \hline
% $Perception$ & $Img \rightarrow Est$ & $Img \cup Est$ \\
% $Controller$ & $Est \rightarrow Cmd$ & $Est \cup Cmd$ \\
% $Dynamics$ & $(Act \times Cmd) \rightarrow Act$ & $Act \cup Cmd$ \\
% $Sensor$ & $(Act \times Env) \rightarrow Img$ & $Act \cup Env \cup Img$ \\
% \hline
% \end{tabular}
% \caption{Component functions and LTS alphabet}
% \label{tab:lts_components}
% \end{table}

\begin{figure}[t]
\centering
  \centering
  \includegraphics[width=0.8\linewidth]{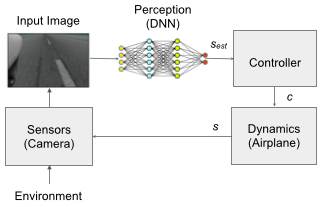}
  \captionof{figure}{Autonomous system using a DNN for perception}
  \label{fig:auto_system}\vspace{-0.7cm}
\end{figure}

Although simple, the type of system we consider resembles (semi-)autonomous mechanisms that are already deployed in practice, such as adaptive cruise controllers and lane-keeping assist systems, which similarly use a DNN for visual perception to provide guidance to the software controlling electrical and mechanical subsystems of modern vehicles. An example of such a system designed for autonomous taxiing of airplanes on taxiways is illustrated in Figure~\ref{fig:auto_system}.  Section~\ref{sec:taxinet} includes a detailed explanation of this system.

We aim to check that the overall system satisfies a {\em safety property} $P$. For the example described in the next section, one such safety property is that 
%an autonomous vehicle does not go off the road.
the airplane does not go off the taxiway, which can be expressed in terms of constraints on the allowed actual system states.
%physical dimensions of the taxiway (see Section~\ref{sec:taxinet}).
%
In order to check this property, one could run many simulations, using,  e.g., XPlane~\cite{xplane}. 
However, simulation alone may not be enough to achieve the high degree of confidence in the  correctness of the system necessary for deployment in a safety-critical setting (e.g., an airplane in our case). We therefore aim to formally verify the property, i.e., we aim to check that $\mathit{System}\models P$ holds.
%is true or produce a counterexample showing otherwise.
%We  write the system as the composition $\mathit{System}=\mathit{Sensors}\parop\mathit{Perception}\parop\mathit{Controller}\parop\mathit{Dynamics}$ and aim to check that $\mathit{System}\models P$ is true. 
 
%To do so we build a discrete model of the (airplane) dynamics and  a discrete-time controller for the system, similar to previous related work \cite{HoffmannTMT07,479140} which also considers discretized control.  
%Treatment of more complex systems with continuous semantics is left for future work.
%\ravi{Perhaps we can also describe how to verify the system using assumptions here. Then, we can point out that either verification or constructing such assumptions with DNN in the loop is infeasible. I can try to write this.}\corina{yes please}
%
Formally verifying $System$ presents serious scalability challenges, even ignoring the learning-enabled aspect, since the conventional components ($Controller$ and $Dynamics$ in our case) can be quite complex; nevertheless they can be tackled with previous techniques, possibly involving abstraction to reduce their state spaces \cite{GiannakopoulouP13}. 
%since a mathematical model of the $Sensor$ can be quite complicated. 
The DNN component makes the scalability problem extremely severe.
%, as we already discussed.
%it can have thousands, millions, or even more parameters, precluding formal verification for most realistic, state-of-the-art networks. Furthermore, the high-dimensional inputs to the DNN come from sensors (themselves complex) and are subject to random %background conditions 
Further, the perturbations from the external world
can not be modeled precisely.

\paragraph{\textbf{Assume-guarantee Reasoning}}

To address the above challenges, we 
%use compositional techniques that decompose the verification of $System$ into the separate verification of its components. In particular, we 
decompose $System$ into two subsystems---$M_1=Controller \parop Dynamics$, i.e., the conventional components,
%$Controller$ and $Dynamics$, 
which can be modeled and analyzed using established model-checking techniques,  on one side, and $M_2=Perception\parop Sensor$, i.e., the perception components, 
%the DNN together with the $Sensor$, 
which are challenging to analyze precisely, on the other side.
%; see illustration in Figure~\ref{fig:comp_verif}. 
%Thus, we decompose $System$ into $M_1=Controller \parop Dynamics$ and $M_2=Perception\parop Sensor$. 
The {\em interface} between $M_1$ and $M_2$ consists of the updates to the actual system states, henceforth called {\em actuals} (performed by the $Dynamics$ component) and to the estimated system states, henceforth called {\em estimates} (performed by the $Perception$ component); let us denote it as $\Sigma_I= \mathit{Act} \cup \mathit{Est}$. We then focus the analysis on $M_1$. 
%This analysis can be further decomposed using similar compositional techniques, to increase scalabiliy. 

Formally checking that system-level property $P$ holds on $M_1$ in isolation does not make too much sense, 
%will likely fail, because 
as $M_1$ is meant to work together with $M_2$ and will not satisfy $P$ by itself (except in very particular cases).
%\ravi{Does $P$ even make sense for $M1$ alone? Dont we have to assume that $\alpha(P)\subseteq \alpha(M_1)$;  i think i assumed that ;  reworded}. 
Assume-guarantee reasoning addresses this problem by checking properties using {\em assumptions} about a component's context (i.e., the rest of the system). 
In the assume-guarantee paradigm, a formula is a triple $\langle A,M,P \rangle$, where $A$ is an assumption, $M$ is a component, and $P$ is a property. The formula is true if whenever $M$ is part of a system that satisfies $A$, the system also guarantees $P$, i.e., $\forall M'$, $M\parop M'\models A \Rightarrow M\parop M'\models P$. For LTSs, this is equivalent to $A\parop M\models P$; $\langle true,M,P \rangle$ is equivalent to $M\models P$. 

Using these triples we can formulate {\rm rules} for compositional, assume-guarantee reasoning.
The simplest such rule allows us to prove that our $System$, composed of $M_1$ and $M_2$, satisfies property $P$, by checking that $M_1$ satisfies $P$ under an assumption $A$ and discharging $A$ on $M_2$: 
%This rule can be represented as follows:

\begin{mathpar}
  \inferrule{\langle A,M_1,P \rangle ~~~ 
  \langle true,M_2,A \rangle}{\langle true,M_1\|M_2,P \rangle} 
\end{mathpar}

% Here $\langle A,M,P \rangle$ denotes an assume-guarantee triple which is true if whenever $M$ is part of a system that satisfies $A$ that system also satisfies $P$.
% We then seek to automatically build an assumption $A$ such that $\langle A, M_1, P\rangle$ holds; one such assumption is the weakest assumption described in Section~\ref{sec:assume_gen}; i.e., by construction $\langle \weakest{\Sigma}, M_1, P\rangle$ is true. \corina{explain} If we can also show that $\langle true, M_2, A\rangle$, then, according to the assume-guarantee rule, it follows that the autonomous system $System=M_1\| M_2$ satisfies the property.

%\ravi{Reworded. Please check corina: rewrote}

%the composition $M\parop M'$ of $M$ with any system $M'$ is guaranteed to satisfy $P$ (i.e., $M\parop M' \models P$) if and only if $M'$ satisfies $A$ (i.e., $M'\models A$). Formally,

%\begin{mathpar}
%  \inferrule{M\parop M' \models P \Leftrightarrow M'\models A}{\langle A,M,P \rangle} 
%\end{mathpar}

We then seek to automatically build an assumption $A$ such that $\langle A, M_1, P\rangle$ holds; one such assumption is the weakest assumption described in Section~\ref{sec:background} for some alphabet $\interalpha\subseteq \Sigma_I$; i.e., by definition $\langle \weakest{\interalpha}, M_1, P\rangle$ is true. If we can also show that $\langle true, M_2, \weakest{\interalpha}\rangle$ is true, then, according to the  rule, it follows that the autonomous system satisfies the property.

Formally checking $\langle true, M_2, \weakest{\interalpha}\rangle$  is infeasible due to the complexity of the DNN and difficulty in mathematically modeling the $Sensor$ component as well as the random environment conditions (as explained before). Instead, we show how the assumption can be leveraged for monitoring (at run-time) the outputs of the DNN, to {\em guarantee} that the overall system satisfies the required property. Furthermore, we show how  automatically generated assumptions can be leveraged for extracting {\em local DNN specifications}, which in turn can be used for training and testing the DNN.

%\subsection
\paragraph{\textbf{$M_1$ Analysis and Assumption Generation}}
%As discussed, we build the weakest assumption  for $M_1$ with respect to property $P$.
%This assumption can be built using the algorithm from ~\cite{ASE02}, which we summarized in Section~\ref{sec:background_assumption}.
%
We first check that $M_1$ does not violate the property assuming {\em perfect perception} (i.e., a simple abstraction that maps each actual to the corresponding estimate). This allows us to uncover and fix all the errors that are due to interactions in the controller and dynamics, independent of errors in perception. 

We then build the weakest assumption  for $M_1$ with respect to property $P$ and interface alphabet $\interalpha\subseteq \Sigma_I$. We use Algorithm~\ref{alg:weakest} which adapts the algorithm from~\cite{ASE02} for our purpose.
\begin{figure}[!t]
 \removelatexerror
  \begin{algorithm2e}[H]
        \KwIn{LTS  model $M$, property $P$, and interface alphabet $\interalpha\subseteq\Sigma_I$}
        \KwOut{Assumption $\weakest{\interalpha}$ for $M$ with respect to $P$, $\interalpha$}
        \Fn{\Fbuildassume{$M,~P,~\interalpha$}}{
        $M'~:=~(M\parop P_{\mathit{err}})\projop{\interalpha}$\;
        $M''~:=~\BackwardErrorPropagation(M')$\;
        $A_{err}^{\interalpha}~:=~\Determinization(M'')$\;
        $\hat{A}_{err}^{\interalpha}~:=~\CompletionWithSink(A_{err}^{\interalpha})$\;
        $\weakest{\interalpha}~:=~\ErrorRemoval(\hat{A}_{err}^{\interalpha})$\;
        \textbf{return} $\weakest{\interalpha}$\;
}
\caption{Computing Weakest Assumption}
\label{alg:weakest}
\end{algorithm2e}\vspace{-0.7cm}
\end{figure}
The function $\Fbuildassume$ has as parameters an LTS model $M$ ($M_1$ in our case), a property $P$, and an interface alphabet $\interalpha$. The first step builds $M\parop P_{\mathit{err}}$ ($P_{err}$ is the complement of $P$) and applies projection with $\interalpha$ to obtain the LTS $M'$. The next step performs backward propagation of $err$ over transitions that are labeled with either $\tau$ or actuals (i.e., actions in $Act$) thus pruning the states where the {\em context} of $M$ can not prevent it from entering the $err$ state. The resulting LTS is further processed with a {\em determinization step} which performs $\tau$ elimination and subset construction (for converting the non-deterministic LTS into a deterministic one). Unlike regular automata algorithms, $\Determinization$ treats sets that contain $err$ as $err$.  
%During subset construction, a state of the determinisic LTS represents a {\em set} of states in the nondeterministic LTS $M'$; if any of these states is $err$, then the whole set becomes $err$ in the deterministic LTS. 
%In this way, if a trace may or may not lead to an error state in $M$ it is considered as an error trace. \ravi{I do not understand the previous sentence. Should it be just "may lead to an error state"}. 
In this way, if performing a sequence of actions from $\Sigma$ does not {\em guarantee} that $M$ is safe, it is considered as an error trace. Subsequently, the resulting deterministic LTS $A_{err}^{\interalpha}$ is {\em completed}  such that every state has an outgoing transition for every action. This is done by adding a special $sink$ state and transitions leading to it. These are missing transitions in $A_{err}^{\interalpha}$ and represent behaviors that are never exercised by $M$; with this completion, they are made into sink behaviors and no restriction is placed on them. 
%
%The resulting LTS, $A_{err,sink}^{\Sigma}$ has both an error state and a sink state, and it is {\em complete}, i.e.,  every state has an outgoing transition for every action. 
The assumption $\weakest{\interalpha}$ is obtained from the complete LTS by removing the $err$ state and all the transitions to it. 

This procedure is similar to the one in~\cite{ASE02} with the difference that the backward error propagation step is performed not only over $\tau$ transitions but also over transitions labeled with actuals. Intuitively, this is because the actuals are updated by $M_1$ (via the $Dynamics$ component in our system) and are only read by $M_2$; thus, the assumption should restrict $M_1$ by blocking the estimates that lead to error but not by refusing to read the actuals. Another difference is that we allow the assumption alphabet to be smaller than $\Sigma_I$ (this is needed as explained later in this section).

By construction,  $\weakest{\Sigma_I}$ captures the traces over  $\Sigma_I^*=(Act\cup Est)^*$ that ensure that $M_1$ does not violate the prescribed safety property, i.e. $\weakest{\Sigma_I}\parop M_1 \models P$ and therefore $\langle \weakest{\Sigma_I}, M_1, P\rangle$.  
Furthermore, $\weakest{\Sigma_I}$ is the weakest assumption, so it does not restrict $M_1$ unnecessarily. 
%\corina{added below; please check; perhaps shorten?}
%\corina{perhaps we need a theorem here to restate this is the weakest assumption, after our modified construction?}
\begin{theorem}
Let $\weakest{\Sigma_I}$ be the LTS computed by $\Fbuildassume(M_1,P,\Sigma_I)$, then 
%$\weakest{\Sigma_I}\parop M_1 \models P$ (and therefore $\langle \weakest{\Sigma_I}, M_1, P\rangle$). Furthermore, 
$\weakest{\Sigma_I}$ is the weakest, i.e., $\forall M_2. M_2 \models \weakest{\Sigma_I}$ iff $M_1 \| M_2 \models P$.  
\end{theorem}
\begin{proof}(Sketch) 
`$\Rightarrow$' similar to \cite{ASE02}.
%Assume $M_1\|M_2\not\models P$. Let $\sigma$ be an error trace in $M_1\|M_2\|P_{err}$; $\sigma$ is a trace in both $M_2$ and $M_1\|P_{err}$. Let $\sigma'$ denote the result of keeping only the actions from $\Sigma_I$ in $\sigma$. $\sigma'$ is an error trace of $M'$ and if it happens to end in an actual, its prefix is in $M''$. Both these traces are in $L(A_{err}^{\interalpha_I})$ and thus can not be in $L(\weakest{\Sigma_I})$  (i.e., the complement), contradicting $M_2 \models \weakest{\Sigma_I}$.\\
`$\Leftarrow$' by contradiction. \\
Assume $M_2 \not\models \weakest{\Sigma_I}$ although $M_1 \| M_2 \models P$. Then there is a trace $\sigma \in L(M_2\projop{\Sigma_I})$ that is also in $L(A_{err}^{\Sigma_I})$. From the construction of the assumption, either (1) $\sigma$ or (2) $\sigma.a$ is in $L((M_1 \| P_{err})\projop{\Sigma_I})$, where $a\in Act$ represents an update to the actuals by $M_1$ and $.$ denotes concatenation. Case (1) is similar to \cite{ASE02}.
%; as $\sigma$ is also in  $L(M_2\projop{\Sigma_I})$, it means $\sigma$ can be used to build a counterexample for $M_1\|M_2\models P$, which is a contradiction. 
Case (2) is new. By construction, $\sigma$ must end in an estimate (due to our special backward error propagation). Furthermore, for our system, actuals and estimates are alternating; thus, $M_2$ must perform a read of the actuals after the estimate and that actual must be $a$; thus $\sigma.a$ is also in  $L(M_2\projop{\Sigma_I})$, and can be used to build a counterexample for $M_1\|M_2\models P$, which is a contradiction.
\end{proof}

%\ravi{should this be $M_2$?}\corina{no, the assumption restricts $M_1$ and must be satisfied by $M_2$; the assumption does not restrict $M_2$;}

%As we shall see, it is also useful to consider a smaller alphabet $\interalpha=Est$ when building the assumptions (as this leads to useful run-time monitors). We note the following property which relates weakest assumptions for different interface alphabets.

%We need the property of being "weakest" in justifying that this is a good runtime monitor since it does not restrict the system unnecessarily. Actually this assumption is only weakest with respect to the alphabet. See Proposition~\ref{prop}.

%\corina{note that if the algorithm for building the assumption returns false with the restricted alphabet, it means that we can not control the system only through monitoring the DNN actions; "false" may be spurious;  how to address: define monitor per initial state}

%\begin{figure}[t]
%\centering
%\begin{minipage}{.5\textwidth}
%  \centering
  %\includegraphics[width=0.9\linewidth]{images/CompVerif.png}
  %\captionof{figure}{Compositional verification}
  %\label{fig:comp_verif}
%\end{minipage}%
%\begin{minipage}{.5\textwidth}
%  \centering
%  \includegraphics[width=0.85\linewidth]{images/RunTime.png}
%  \captionof{figure}{Autonomous system with run-time monitor}
%  \label{fig:run-time}
%\end{minipage}
%\end{figure}

%\subsection
\paragraph{\textbf{Uses of Assumptions}} %$\weakest{\interalpha}$ and %$\weakest{\Sigma_I}$}}
\label{sec:assume_uses}
%\ravi{We should maybe use a different notation instead of $\weakest{\Sigma}$ when referring to the subset of $\Sigma_I$}

\noindent\textbf{\\$\weakest{Est}$ for Run-time Monitoring.} The assumptions $\weakest{\interalpha_I}$ can potentially be used as a monitor deployed at run-time to ensure that the autonomous system guarantees the desired safety properties. One difficulty is that $\Sigma_I$ refers to labels in both $Est$ and $Act$ which represent the estimated and actual values of the system states, respectively.  However the autonomous system may not have access to the actual system state---the very reason it uses a DNN is to get the estimated values of the system state.

While in some cases, it may be possible to  get the actual values through alternative means, e.g., through other sensors,  we can also set the alphabet of the assumption to be only in terms of the estimates received from the DNN, i.e., $\interalpha=Est$, and build a run-time monitor solely based on $\weakest{Est}$. 

Since $\weakest{Est}$ is modeled only in terms of the $Est$ alphabet, it follows that it can be deployed as a run-time monitor on the outputs of a DNN that is used by the autonomous system.

%As we shall see, it is also useful to consider a smaller alphabet $\interalpha=Est$ when building the assumptions (as this leads to useful run-time monitors). 
We note the following property which relates weakest assumptions for different interface alphabets.

\begin{proposition}\label{prop}
%Assume LTSs $M$, property $P$, and interface alphabet $\Sigma_I\subseteq \alpha M$. Assume also $\Sigma$, $\Sigma$ such that $\Sigma \subseteq \Sigma' \subseteq \Sigma_I$. Then, $L(A_{w,\Sigma})\subseteq L(A_{w,\Sigma'}) \subseteq L(A_{w,\Sigma_I})$.
Given LTS $M$, property $P$, and interface alphabet $\Sigma_I\subseteq \alpha(M)$,   $\forall \interalpha$, $\interalpha'$.  $\interalpha \subseteq \interalpha' \subseteq \Sigma_I \Rightarrow L(\weakest{\interalpha})\subseteq L(\weakest{\interalpha'}) \subseteq L(\weakest{\Sigma_I})$.%\ravi{do we need to add projection?}
\end{proposition}
%\begin{proof}
%\corina{to prove?}
%(Sketch) 
The intuition is that projection with a  smaller alphabet results in more behaviours in $M$ leading also to more error behaviours; consequently, the corresponding assumptions are more restrictive (contain less behaviours). Conversely, by adding actions to the interface alphabet, the corresponding assumption becomes weaker, more permissive (see also~\cite{alphatacas}). 
%\end{proof}

Thus, $\weakest{\interalpha}$ is only weakest with respect to a particular interface alphabet $\interalpha$. Moreover, $L(\weakest{\interalpha})$ can be empty if no assumption with alphabet $\interalpha$ can ensure that the system $M$ satisfies property $P$. In section~\ref{sec:eval}, we will describe an experiment that aims to quantify how permissive the assumption $\weakest{Est}$ is in the context of a system that uses a realistic DNN  for perception.

\noindent\textbf{$\weakest{Est}$ for Testing and Training DNNs.} 
The extracted assumptions over alphabet $Est$ can also be used for testing a candidate DNN to ensure that it follows the behaviour prescribed by the assumption. For many autonomous systems (see e.g., the airplane taxiing application in section~\ref{sec:taxinet}), the perception DNN is trained and tested based on images obtained from simulations which naturally come in a sequence, and therefore, can be easily checked against the assumption by evaluating if the sequence of DNN predictions represents a trace in $L(\weakest{Est})$.  The assumption can also be used during the training of the DNN as a specification of desired output for unlabeled real data, thus reducing the burden of manually labeling the input images. We leave %the exploration of 
these directions for future work.

\noindent\textbf{$\weakest{\Sigma_I}$ for Synthesizing Local Specifications.}
% - For every state $s$ in $A_{err}^{\Sigma_I}$:\\
% -- if $s$ has outgoing transitions leading to error:\\
% --- Let $E=\{e | (s,e,err)\in\delta\}$ \\
% --- Let $E'=Est-E$ (allowed estimates);\\
% --- For every incoming transition to $s$, $(s',a,s)$:\\
% --- Let $c_a$ define the constraint on actuals corresponding to $a$, and $c_{e_i}$ define the constraints on estimates corresponding to each $e_i\in E'$;\\
% --- Output local specification $c_a \Rightarrow  (\bigvee c_{e_i})$.
%
\begin{figure}[!t]
 \removelatexerror
  \begin{algorithm2e}[H]
        \KwIn{$A^{\Sigma_I}_{err} = (Q,\Sigma_I,\delta,q_0)$}
        \KwOut{Local specifications $\Phi$}
        \Fn{\FSynthSpec{$A^{\Sigma_I}_{err}$}}{
        $\Phi~:=~\{\}$\;
        \ForEach{$q \in Q$}{
            \If{$\exists a. (q,a,err) \in \delta$}{
                $E~:=~\{a \mid (q,a,err) \in \delta\}$\;
                $E'~:=~Est - E$\;
                \ForEach{$(q',a',q) \in \delta$}{
                    $\phi~:=~ (s=a') \Rightarrow \bigvee_{a \in E'}(s_{est}=a)$\;
                    $\Phi~:=~\Phi \cup \phi$\;
                }
            }
        }
        \textbf{return} $\Phi$\;
}
\caption{Synthesizing Local Specifications}
\label{alg:synthesize_local}
\end{algorithm2e}\vspace{-0.7cm}
\end{figure}
We also propose to analyze the (complement of the) weakest assumption generated over the full interface alphabet $\Sigma_I = Act \cup Est$ to synthesize local, non-temporal specifications for the DNN. These specifications can be used as formal {\em documentation} of the expected DNN behavior; furthermore, they can be leveraged to train and test the DNN. Unlike the temporal LTS assumptions, evaluating the DNN with respect to local specifications does not require sequential data, making them more natural to use when evaluating DNNs.

Algorithm~\ref{alg:synthesize_local} describes a procedure for synthesizing such local specifications. The input to the algorithm is {\em the complement} of the assumption,  i.e., the output of line 4 in Algorithm~\ref{alg:weakest}, which encodes the error traces of the assumption. We aim to extract local specifications from the error transitions.
%\ravi{Decide whether to have $sink$ in the notation or not.} 
We first note that in $A^{\Sigma_I}_{err}$, only transitions corresponding to estimates (i.e., labeled with elements from $Est$) can lead to the $err$ state; this is due to our special error propagation. 
%In other words, transitions in $A^{\Sigma_I}_{err}$ corresponding to actuals (i.e., labeled with elements from $Act$) are always followed by an $Est$-labeled transition.
%Intuitively, this happens because in $M_1$, when an error state is entered, the actuals are no longer updated. 
Furthermore, actuals and estimates are alternating, due to the nature of the system. %(an estimate is always preceded by an actual). 
%Intuitively, this happens because for a safety property $P$ that restricts the allowed actuals, the subsystem $M_1 = Controller \parop Dynamics$ can reach the $err$ state only after the $Controller$ reads the estimates and issues a command. \ravi{In our models, a "bad" command directly causes evolution to the $err$ state without an update in the actuals. Hence, the last action before evolution to $err$ is in $Est$.} If the command does not cause the actuals to evolve to $err$, then a transition corresponding to an actual is necessarily followed by a transition corresponding to an estimate. 
Algorithm~\ref{alg:synthesize_local} exploits this structure in $A^{\Sigma_I}_{err}$ to synthesize local specifications. 
%\corina{added this}We envision that this algorithm can be generalized, to make it applicable to other systems that do not necessarily follow this structure.

For each state $q$ in $A^{\Sigma_I}_{err}$ (line 3) that can directly transition to the $err$ state (line 4), the algorithm first collects all the actions $a$ that lead to $err$ (line 5). As described earlier, these actions belong to  $Est$. Next, for each incoming transition to $q$ (line 7), we construct a local specification $\phi$ (line 8). Each incoming transition to $q$  corresponds to an action $a' \in Act$ as described earlier. The synthesized local specification expresses that for an actual system state $s$ with value $a'$, the corresponding estimated system state ($s_{est}$) should have a value in $E'$ to avoid $err$. 
%Note that these local specifications tolerate some error in the estimation, as they do not strictly prescribe that the estimated values (i.e., the outputs of the DNN) should be the same as the actual values (i.e., the ground truth for the DNN).

We can argue that if $M_2 = Perception \parop Sensor$ satisfies these local specifications then it also satisfies the assumption (proof by contradiction). Intuitively, this is true because the local specifications place stronger requirements on $M_2$ compared with the assumption $\weakest{\Sigma_I}$.

\begin{theorem}
For $A^{\Sigma_I}_{err}$ and $M_2= Perception\parop Sensor$, if $M_2$ satisfies local specifications $\Phi = \FSynthSpec(A^{\Sigma_I}_{err})$, then $\langle true, M_2, \weakest{\Sigma_I}\rangle$ holds.
\end{theorem}

\begin{proof} 
(Sketch) Assume that $\langle true, M_2, \weakest{\Sigma_I} \rangle$ does not hold, i.e., there is a counterexample trace $\sigma$ of $M_2$ that violates the assumption; this is a trace in $A^{\Sigma_I}_{err}$. Due to our error propagation, it must be the case that the last action in this trace is an estimate $s_{est} \in Est$. Let $q_i$ be the state in $\weakest{\Sigma_I}$ that is reached by simulating $\sigma$ on $\weakest{\Sigma_I}$ prior  to the last, violating estimate $s_{est}$. Since $M_2$ satisfies the local specification for $q_i$ it means there can be no $s_{est}$ leading to $err$ from $q_i$, a contradiction.
\end{proof}

Furthermore, if $\langle true, M_2, \weakest{\Sigma_I} \rangle$ holds, then,  according to the assume-guarantee reasoning rule, it follows that the $System = M_1 \parop M_2$ satisfies the required properties.

%\corina{talk about the use of these properties}

While it may be infeasible to {\em formally prove} such properties for $M_2$,
%due to scalability issues, difficulty of mathematically modeling $Sensor$, as well as challenges in characterizing the random external environment $Env$,
%not being able to encode them in a checkable form, 
 these local specifications can be used instead for testing and even training the DNN. Given an image $img$ labeled with the underlying actual $a'$ (i.e., the $Sensor$ produces $img$ when the actual state is $a'$), we can test the DNN against the local specification $s=a' \Rightarrow \bigvee_{a \in E'}(s_{est}=a)$, by checking if the DNN prediction on $img$ satisfies the consequent of the specification. Compared with  the standard DNN testing objective that checks if the state estimated by the DNN \emph{matches} the underlying actual system state, our local specifications yield a relaxed testing objective.
%(i.e., the ground truth), with a relaxed testing objective. 
Similarly, these specifications can also be used during training to relax the training objective. Instead of requiring the DNN to predict the actual system state from the image, under the relaxed objective, any prediction that satisfies the corresponding local specification is acceptable. Such a relaxed objective could potentially lead to a better DNN due to the increased flexibility afforded to the training process, but we leave the exploration of this direction for future work.

\section{The TaxiNet System}
\label{sec:taxinet}
 
%In this section, we present a detailed discussion of the application our compositional approach to a case study. 
We present a case study applying our compositional approach to 
%a realistic autonomous system. 
%
%\subsection{The Taxinet Network}
%In particular, we analyse the design models for an 
an experimental autonomous system for center-line tracking of airplanes on airport taxiways~\cite{ACT1,pasareanu2023closedloop}. The system  uses a DNN called TaxiNet for perception. TaxiNet is a regression model with  24 layers including five convolution layers, and three dense layers (with 100/50/10 ELU~\cite{clevert2015fast} neurons) before the output layer. TaxiNet is designed to take a picture of the taxiway as input and return the plane’s position with respect to the center-line on the taxiway. It returns two outputs; cross track error ($\vcte$), which is the distance in meters of the plane from the center-line and heading error ($\vhe$), which is the angle in degrees of the plane with respect to the center-line. These outputs are fed to a controller which in turn manoeuvres the plane such that it remains close to the center of the taxiway. This forms a closed-loop system where the perception network continuously receives images as the plane moves on the taxiway.

%TaxiNet has been trained and tested using the X-Plane simulator~\cite{xplane}. 
%Though center-lines on the pavement of airport taxiways and taxiways have standardized shapes and colors, their visibility maybe poor for a number of reasons, including skid marks on the taxiways, poor lighting conditions, and bad weather. 
The architecture of the system is the same as in  Figure~\ref{fig:auto_system}.
For this application, state $s$ captures the position of the airplane on the surface in terms of $\vcte$ and $\vhe$ values.

%\subsection
\paragraph{\textbf{Safety Properties}}
We aim to check that the system satisfies two safety properties, as provided by the industry partner. The properties specify conditions for safe operation in terms of allowed $\vcte$ and $\vhe$ values for the airplane by using taxiway dimensions.  The first property states that the airplane shall never leave the taxiway (i.e., $|\vcte|\leq8$ meters). The second property states that the airplane shall never turn more than a prescribed degree (i.e., $|\vhe|\leq35$ degrees), as it would be difficult to manoeuvre the airplane from that position.
%
%Note that, by construction, the DNN's outputs do not violate the safety properties \corina{expand}; 
Note that the DNN output values are normalized to be in the safe range;
however, this does not preclude the overall system from reaching an error state.

%While we could express both as property LTSs, for simplicity, we encode them here as error states in the code of the Dynamics model, where an error for either $\vcte$ or $\vhe$ indicates that the airplane is off the taxiway or turned more than the prescribed angle, respectively. LTSA can then check reachability of these error states automatically.

%\subsection{Component Modeling}
%The code for the controller and the dynamics is as follows:
\begin{figure}[!t]
\centering
\input{code/code_ltsa.tex}
\caption{$Controller$ and $Dynamics$ in the 
process-algebra style 
FSP language for the LTSA tool.
%(see Appendix~\ref{sec:code_ltsa_pessimistic} for complete code).
%In FSP, 
Identifiers starting with lowercase/uppercase letters denote labels/processes (states in the underlying LTS), respectively; $\rightarrow$ denotes labeled transitions between states.
%in the underlying LTS,
%(for instance, $\sigma\rightarrow \sigma'\rightarrow P$ is equivalent to the process algebra term $\sigma\cdot\sigma'\cdot P$ denoting a pair of transitions $(q,\sigma,q')$ and $(q',\sigma',q'')$ where $q,q',q''$ are states and $\sigma,\sigma'$ are actions in the LTS). 
%$\mid$ denotes a choice operator. 
Both labels and processes can be indexed.
%; e.g., {\tt estimate[c:CTERange] [h:HERange]} appears internally as {\tt estimate[0][0], estimate[0][0], estimate[0][2]}, etc. 
Conditionals have the usual meaning.
%and are used to increase expressivity of the language.
%\corina{maybe move this to appendix}
\vspace{-0.7cm}}
\label{fig:taxinet-lts}
\end{figure}

\paragraph{\textbf{Component Modeling}}
We build a discrete-state model of $M_1=Controller\parop Dynamics$ as an LTS. We assume a discrete-event controller and a discrete model of the aircraft dynamics.
%The $Controller$ and the $Dynamics$ reason about the actual and estimated values of the system state via discrete ranges {\tt CTERange} and {\tt HERange} which correspond to conditions on the outputs of the DNN, $\vcte \in [-8,8]$ and $\vhe \in [-35,35]$. 
%We use a fixed discretization for {\tt HERange} and experiment with discretizations at different granularities for {\tt CTERange}, as defined by a parameter {\tt MaxCTE}. For instance, when {\tt MaxCTE=2} the discretization divides the $\vcte$ range evenly into three intervals as follows:
The $Controller$ and the $Dynamics$ operate over discretized actual and estimated values of the system state. 
We use a fixed discretization for $\vhe$ and experiment with discretizations at different granularities for $\vcte$, as defined by a parameter \fspsyn{MaxCTE}. For instance, when \fspsyn{MaxCTE=2}, the discretization divides the $\vcte$ and $\vhe$ as follows.
%evenly into three intervals; the $\vhe$ range is divided evenly as well:
%\corina{to change as now we expresses a parameterized discretization}
\noindent\begin{minipage}{0.5\linewidth}
\begin{equation*}
\resizebox{0.9\hsize}{!}{
  $\vctec =
  \setlength{\arraycolsep}{0pt}
  \renewcommand{\arraystretch}{1}
  \left\{\begin{array}{l @{~} l c} 
        0            & \text{if}~\vcte \in [8, -2.7) %& ({\tt LeftMargin})
        \\
        
        1            & \text{if}~\vcte \in [-2.7, 2.7] %& ({\tt Center})
        \\
       
        2            & \text{if}~\vcte \in  (2.7, 8] %& ({\tt RightMargin})
  \end{array}\right.$}
\end{equation*}
\end{minipage}%
~
\begin{minipage}{0.5\linewidth}
\begin{equation*}
\resizebox{0.9\hsize}{!}{
  $\vhec =
  \setlength{\arraycolsep}{0pt}
  \renewcommand{\arraystretch}{1}
  \left\{\begin{array}{l @{~} l c} 
        1           & \text{if}~\vhe\in[-35, -11.67)%& ({\tt NegAngle})
        \\
        0           & \text{if}~\vhe\in[-11.67, 11.66]%& ({\tt NoAngle})
        \\
        2           & \text{if}~\vhe\in(11.66, 35.0]%& ({\tt PosAngle})
  \end{array}\right.$}
\end{equation*}
\end{minipage}

\noindent For simplicity, we use $\vcte$ and $\vhe$ to denote both the discrete and continuous versions in other parts of the paper (with meaning clear from context).

%\todo{Make sure notation is used consistently. For instance, earlier $err$ is a state but here $err$ is an action. Also check $est$ and $act$.}
Figure~\ref{fig:taxinet-lts} gives a description of the $Controller$ and $Dynamics$ components. We use \fspsyn{act[cte][he]} to denote actual states $s$ in the $Act$ alphabet and \fspsyn{est[cte][he]} to denote the estimated states $s_{est}$ in $Est$. While we could express the safety properties as property LTSs, for simplicity, we encode them here as the \fspsyn{ERROR} states in the LTS of the $Dynamics$ component, where an error for either $\vcte$ or $\vhe$ indicates that the airplane is off the taxiway or turned more than the prescribed angle, respectively. 
%The \fspsyn{ERROR} states, reached by transitions labeled \fspsyn{err}, correspond to the property $err$ state described in Section~\ref{sec:background}. 
Tools for analyzing LTSs, such as LTSA~\cite{ltsa}, can check reachability of error states automatically. 

% We present the LTSs for these components below (using LTSA's FSP input language \cite{fsp}).
% %; the complete code is in the \corina{Appendix}.
% In FSP, identifiers starting with lowercase letters denote labels while identifiers starting with uppercase letters denote processes (states in the underlying LTS); symbol $\rightarrow$ models labeled transitions between states in the underlying LTS, while $\mid$ denotes a choice operator. Both labels and processes can be indexed.
% %; e.g., {\tt estimate[c:CTERange] [h:HERange]} appears internally as {\tt estimate[0][0], estimate[0][0], estimate[0][2]}, etc. 
% Conditionals have the usual meaning and are used to increase the expressive power of the language.

The $Controller$ reads the estimates via ${\tt est}$-labeled transitions. The $Controller$ can take three possible actions to steer the airplane---\fspsyn{GoStraight}, \fspsyn{TurnLeft}, and \fspsyn{TurnRight} (denoted by \fspsyn{cmd[0]}, \fspsyn{cmd[1]}, and \fspsyn{cmd[2]} respectively).
The $Dynamics$ updates the system state, via ${\tt act}$-labeled transitions; the initial state is \fspsyn{act[1][0]}. Action \fspsyn{turn} is meant to synchronize the $Controller$ and the $Dynamics$, to ensure that the estimates happen after each system update.
%\ravi{Not sure if I understand the need for \fspsyn{turn}. Won't they synchronize on the $cmd$ actions?}

%\subsection{Abstractions for Perception}

%As mentioned, we use two abstractions (optimistic vs. pessimistic) for the perception and the camera components of the system. 
%\begin{lstlisting}[language={Prism}, rulesepcolor=\color{black}, rulecolor=\color{black}, breaklines=true, breakatwhitespace=true,frame=single,
%basicstyle=\scriptsize
%\color{black}]
%// "optimistic" view
%Perception_best = S0[0][0],
%S0[c0:CTERange][h0:HERange] =
% (actual[c0][h0] -> estimate[c0][h0] -> S1),
%S1 = (next_actual[c0:CTERange][h0:HERange] -> S0[c0][h0]).
%\end{lstlisting}

%\begin{lstlisting}[language={Prism}, rulesepcolor=\color{black}, rulecolor=\color{black}, breaklines=true, breakatwhitespace=true,frame=single,
%basicstyle=\scriptsize
%\color{black}]
%// "pessimistic" view
%Perception_worst = S0[0][0]),
%S0[c0:CTERange][h0:HERange] = (actual[c0][h0] -> estimate[c1:CTERange][h1:HERange] -> S1),
%S1 = (next_actual[c0:CTERange][h0:HERange] -> S0[c0][h0]).
%\end{lstlisting}

%In the optimistic view, an actual state $(c_0,h_0)$  maps precisely to the estimated state $(c_0,h_0)$, modeling the fact that the perception is perfectly accurate. In the pessimistic view, an actual state $(c_0,h_0)$ maps to any other estimated state $(c_1,h_1)$, ranging over 
%{\tt CTERange} $\times$ {\tt HERange}, modeling that the perception can make {\em any} mistakes in the estimation.

We analyze $M_1=Controller\parop Dynamics$ as an {\em open} system; in $M_1$ the estimates can take {\em any} values (see transition labeled \fspsyn{ est[cte:CTERange][he:HERange]} in the $Controller$), irrespective of the values of the actuals. Thus, we implicitly take a {\em pessimistic view} of the $Perception$ DNN and assume the worst-case---the estimates can be arbitrarily wrong---for its behavior. It may be that a well-trained DNN with high test accuracy may perform much better in practice than this worst-case scenario. However, it is well known that even highly trained, high performant DNNs are vulnerable to adversarial attacks, natural and other perturbations as well as to distribution shifts which may significantly degrade their performance. We seek to derive strong guarantees for the safety of the overall system even in such adversarial conditions, hence our {\em pessimistic} approach.

Note also that when using an optimistic $Perception$ component, LTSA reports no errors, meaning that the system is safe assuming no errors in the perception.
%; see Appendix~\ref{sec:code_ltsa_optimistic}.

%\subsection
\paragraph{\textbf{Assumption Generation}}

%We illustrate the analysis for the simple case of {\tt CTEMax=2}, i.e., there are only three possible discrete values for \vcte. 

%When checking $M_1$ (for {\tt MaxCTE=2}), LTSA reports that an error is reachable. This is not surprising since the estimated values are completely unconstrained.  Specifically, LTSA reports the following counterexample:

%\begin{verbatim}
%  start.1.0
%  turn
%  estimate.0.0
%  action.2
%  next_actual.2.2
%  actual.2.2
%  turn
%  estimate.0.0
%  action.2
%  err   
%\end{verbatim}

%The counterexample can be interpreted as follows. The system starts with actuals [\vcte][\vhe] set to [1][0] (the airplane is on the centerline, no angle). But the estimate is [0][0] (indicating that the perception network made a mistake when estimating  the value for \vcte). Based on the estimates, the controller issues action '2' (turn right) to correct the estimated \vcte. Based on this action and the actuals, the Dynamics updates the next actuals to be [2][2] (the position of the airplane is now center-right at a positive angle).
%The perception component again makes a mistake and estimates [0][0] instead of the correct [2]2]. Based on this information, the controller again tries to steer the controller to the right, leading to the airplane violating both safety requirements (i.e., going off the runway and being at more than the prescribed angle).

We build an assumption, using the procedure described in Algorithm~\ref{alg:weakest}, that {\em restricts} $M_1$ in such a way that it satisfies the safety properties.
%no longer exhibits counterexample traces such as the above. 
At the same time, the assumption does not restrict $M_1$ unnecessarily, i.e., it allows $M_1$ to operate normally, as long as it can be prevented (via parallel composition with the assumption) from entering the error states. 
%LTSA uses Algorithm~\ref{alg:weakest} for computing the assumption.

For the assumption alphabet, we consider $\Sigma_I$ = \fspsyn{act[CTERange] [HERange]} $\cup$ \fspsyn{est[CTERange][HERange]} which consists of actual and estimated values exchanged between $M_1$ and $M_2$.
%(see Figure~\ref{fig:comp_verif}).
 As mentioned in Section~\ref{sec:verif}, while the resulting assumption $\weakest{\Sigma_I}$ can be leveraged for synthesizing local specifications, using it as run-time monitor can be difficult since the actual values of the system state may not be available at run-time with the system accessing the external world only through the values that are estimated by the DNN.
%The resulting assumption $\weakest{\Sigma_I}$ may be useful for training and testing a DNN. 
%$\weakest{\Sigma_I}$ may also be useful for implementing a run-time safety monitor, however one difficulty would be that the actual values of the system state may not be available at run-time, as the system has access to the external world only through the values that are estimated by the DNN. 
We therefore define a second alphabet, $\interalpha$=\fspsyn{est[CTERange][HERange]}, which consists of only the estimated values, and build a second assumption $\weakest{Est}$.
%(here, we used the LTSA shortcut notation for alphabets \ravi{not sure what shortcut notation is referring to}). 
We describe these two assumptions in more detail below.

\begin{figure}[t]
\centering
 \includegraphics[width=1.1\linewidth]
 {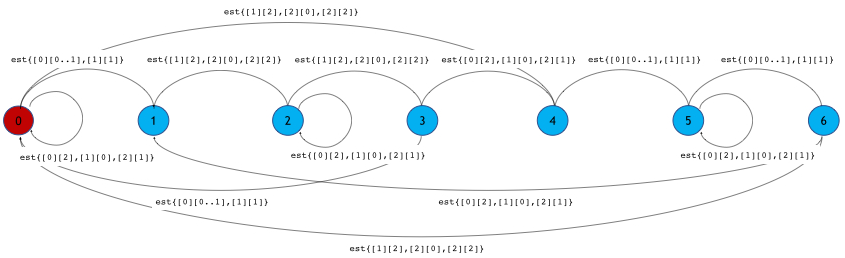}
 %{images/assume_small.png}
 %{AssumptionClean.png}
 %\hspace{-5cm}
 %\includegraphics[width=0.8\linewidth]{Assumption_TaxiNet.pdf}
 %\vspace{-5cm}
 \captionof{figure}{Assumption $\weakest{Est}$ for TaxiNet when \fspsyn{MaxCTE = 2}.}
  %\captionof{figure}{Error LTS for generated assumption. Transitions marked with a cross indicate erroneous behaviour going to error state (-1). Remaining transitions represent the weakest assumptions for safe operation. }
  \label{fig:assumption}\vspace{-0.5cm}
\end{figure}

\noindent\textbf{Assumption $\weakest{Est}$.}
Figure~\ref{fig:assumption} shows the assumption that was generated for the alphabet $\interalpha$ consisting of only the estimated values (\fspsyn{est[CTERange][HERange]}).

%Let us analyze the counterexample above, and consider from it only the actions in {\tt estimate[CTERange][HERange]}.

%Intuitively, trace: 'estimate[0][1], estimate[0][1]' can not be in $L(A_{w,\Sigma})$ since the counterexample shows that if the perception gives two estimates of [0][1] in a row, the system $M$ will violate the safety properties. However, trace 'estimate[0][1]' should be allowed in $L(A_{w,\Sigma})$, since after only one estimate of [0][1] from the perception it is not the case that the system goes to error.

%To build the assumption, we invoked the algorithm in~\cite{ASE02} (summarized in Section~\ref{sec:background_assumption}) 
%on {\tt M1\verb|@|\{estimate[CTERange][HERange]\} } (\verb|@| corresponds to $\projop$ in LTSA). 

%Figure~\ref{fig:assumption} shows error LTS for the obtained assumption; we obtained the error LTS for the assumption by using the {\tt property} keyword in LTSA 
%The figure was automatically generated by LTSA, but we highlighted some states and transitions to ease the readability.

In the figure, each circle represents a state. The initial state (0) is shown in red. %and -1 is the error state of the LTS. The transitions going into the error state are marked with a cross. The remaining transitions represent the weakest assumptions for $M$.
%(enlisted below).
Let us look at some of the transitions in detail. The initial state has a transition leading back to itself with labels \fspsyn{est[0][2],est[1][0],est[2][1]}. 
%This indicates that if the DNN keeps estimating for \vcte and \vhe  either [LeftMargin][PosAngle] ([0,2]) or [Center] [NoAngle] ([1][0]) or [RightMargin][NegAngle] ([2][1]), then the system continues to remain safe, regardless of the actuals.
This indicates that if the DNN keeps estimating either \fspsyn{[0][2]} or \fspsyn{[1][0]} or \fspsyn{[2][1]} for $\vcte$ and $\vhe$, then the system continues to remain safe, regardless of the actuals. 
Intuitively, this is true because the system starts in initial actual state \fspsyn{[1][0]} and 
all three estimates (\fspsyn{[0][2], [1][0], [2][1]}) lead to the same action issued by the controller, which is \fspsyn{GoStraight}, ensuring that the system keeps following the straight line, never going to error.  

%Furthermore, let us go back to our counterexample and consider from it only the estimates. When the first estimate is  [0][0], the assumption transitions from the initial state to state 1. However, if the next estimate is also the same, as in the counterexample, this is not allowed in the assumption (i.e., there is no outgoing transition from state 1 with label [0][0]). Thus, the assumption  is able to {\em block} this error behavior.

%then the plane can go off the runway (refer transition from state 1 to -1). For instance, consider an estimate of [1][0] (left,heading straight), the corresponding action by the controller would be a = 2(turn right). This in turn would make the plane head right and move to the right of the center-line, making the actual cte and he as [2][2]. However, if the estimate is again [1][0], the action of turn right, would move the plane out of the runway (he > 35\degree).

%Note that just allowing the estimates $\{[0][0],[1][2],[2][1],[3][2],[4][1]\}$, would keep the system in the initial state and satisfy the safety property. However, this would restrict the possible behaviors of the model. The transitions covered in the assumptions are the weakest or the most permissive while satisfying the safety property. 

The assumption  
%captures precisely the allowed DNN output sequences that ensure that the overall system does not violate the prescribed safety properties. Thus it 
can be seen as a {\em temporal specification} of the DNN behaviour which was derived automatically from the  behaviour of $M_1$ with respect to the desired safety properties. 
%\corina{comment on the assumption; relationship with theorem; specific to a scenario; initial state}
%The transitions sinking into the error state represent the DNN output sequences that violate the safety property and can be used to build runtime safety monitors. 

%\corina{perhaps say something about building an assumption per initial state; }

\noindent\textbf{Assumption $\weakest{\Sigma_I}$.}
%\corina{not sure if we should show it}
The code for $A^{\Sigma_I}_{err}$, generated for the purpose of synthesizing local specifications using the alphabet $\Sigma_I$ with both actual and estimated values (\fspsyn{act[CTERange][HERange], est[CTERange][HERange]}), is shown in Figure~\ref{fig:err_assumption}.
Recall that this code is the result of step 4 in Algorithm~\ref{alg:weakest}; thus it encodes the {\em error behaviour} of the perception in terms of estimated and ground-truth (actual) output values for the DNN. 
%of $M_1$, i.e., \ravi{changed} traces in terms of estimated and actual state values that lead to error states. 
%The assumption is in terms of {\em sequences} describing the evolution (in time) of the DNN.

\begin{figure}[!t]
\centering
\input{code/code_assumption.tex}
\caption{$A^{\Sigma_I}_{err}$ for TaxiNet when $\Sigma_I = Est \cup Act$ and \fspsyn{MaxCTE=2}. We show it in textual form for readability.}
\label{fig:err_assumption}\vspace{-0.7cm}
\end{figure}

% We first note that the assumption never puts restrictions on the actual values; i.e., only estimate-labeled transitions can lead to the ERROR state. Intuitively, this happens because the system can not enter an error state after an update of an actual value (there is always an estimate after an actual); so the errors can only happen due to wrong estimations; consequently, the assumption only restricts the estimated values. Furthermore, actuals and estimates are alternating, due to the nature of the system (an estimate is always preceded by an actual). We aim to extract local specifications from the error transitions.

%Let us analyze this assumption to extract local non-temporal specifications of the neural network. 
%these specifications can be used to better test a neural network. 
Let us consider how Algorithm~\ref{alg:synthesize_local} synthesizes local, non-temporal specifications using this code. 
For instance, in state \fspsyn{Q3}, estimates \fspsyn{\{[0][0..2], [1][0..1], [2][1]\}} lead to error, thus only estimates \fspsyn{[1][2]}, \fspsyn{[2][0]}, and \fspsyn{[2][2]} are safe. Furthermore, \fspsyn{Q3} is reached (from \fspsyn{Q1}) when the actual state is \fspsyn{[2][2]}.
Following similar reasoning, in state \fspsyn{Q5}, estimates \fspsyn{[0][2]}, \fspsyn{[1][0]}, \fspsyn{[1][2]}, \fspsyn{[2][0..2]}  are safe (since estimates \fspsyn{\{[0][0..1], [1][1]\}} lead to error) and \fspsyn{Q5} is reached when actual is \fspsyn{[2][0]}. Similar patterns can be observed for \fspsyn{Q7}, \fspsyn{Q10}, \fspsyn{Q12}, and \fspsyn{Q14}.

From \fspsyn{Q3}, we can infer the following local specification for the DNN:
%\begin{equation*}
%\begin{aligned}
$ (\vcte^*=2~\wedge~\vhe^*=2) \Rightarrow \big((\vcte=1~\wedge~ \vhe=2)~\vee 
 (\vcte=2~\wedge~\vhe=0)~\vee~(\vcte=2~\wedge~\vhe=2)).$
%\end{aligned}
%\end{equation*}
Here '*' denotes actual state values. 
%(i.e., the ground truth for the DNN). 
This specification gets translated back to the original, continuous DNN outputs as follows:
%\begin{equation*}
%\begin{aligned}
$ (\vcte^* \in [2.7,8) \wedge \vhe^* \in  (11.66, 35.0]) \Rightarrow 
 \big((\vcte\in[-2.7,2.7]~
\wedge~\vhe\in(11.66, 35.0])~\vee
 (\vcte\in [2.7, 8)~\wedge~\vhe\in[-11.67,11.66])~\vee
 (\vcte \in [2.7, 8)~\wedge~\vhe \in  (11.66, 35.0])).$
%\end{aligned}
%\end{equation*}
This specification can be interpreted as follows. For an input image that has ground truth $\vcte^* \in [2.7,8) \wedge \vhe^* \in  (11.66, 35.0]$, the output of the DNN on that image should satisfy $(\vcte\in[-2.7,2.7]
\wedge \vhe\in(11.66, 35.0])\vee
(\vcte\in [2.7, 8) \wedge \vhe\in[-11.67,11.66])\vee
(\vcte \in [2.7, 8) \wedge \vhe \in  (11.66, 35.0])$.

%We note that this local specification tolerates some error in the estimation. Unlike in the standard DNN testing objective, which requires that for every input image, the output of the DNN should match the ground truth, our local 
Thus, the specification tolerates some DNN output values that are different than the ground truth, as they do not affect the safety of the overall system.  The industry partner is using these specifications to help elicit DNN requirements, 
%and also to retrain and test the TaxiNet.
%\corina{perhaps say something about the industry partner; he found these properties useful?}
%The specifications helps to identify and refine the performance requirements for the DNN components, 
which are always a challenge in learning-enabled system development. The specifications can also be used to support sensitivity analysis of the DNN. The requirements and sensitivity analysis are important contributors in the assurance of learning-enabled safety-critical systems.
%\corina{to explain a bit more}

\section{Evaluation}
\label{sec:eval}
%In this section, we use the TaxiNet case study to evaluate the effect of discretization granularity on the size of the generated assumptions and on the scalability of the assumption generation algorithm. We also evaluate the use of assumptions as run-time safety monitors. We leave the evaluation of local specifications for future work.

%\subsection
\paragraph{\textbf{Assumptions for Increasing Alphabet Sizes}}
\begin{table}
\centering
\begin{tabular}
{|c|c|c|c|c|c|}\hline
\fspsyn{MaxCTE}&Assumption&$M_1$&Time&Memory\\
&size&size&(seconds)&(KB)\\
\hline
2   &   7   & 99 & 0.079  & 9799\\
4   &   13  & 261 & 0.126 & 10556\\
6   &   19  & 495 & 0.098 & 9926\\
14  &   43  & 2151 & 0.143 & 13324\\
30  &   91  & 8919 & 0.397 & 31056\\
50  &   151 & 23859 & 2.919 & 45225\\
100 &   301 & 92709 & 81.529 & 132418\\
\hline
\end{tabular}
\caption{Effect of discretization granularity on assumptions.}
\label{tab:assumption_results}\vspace{-0.7cm}
\end{table}

%As our model is parameterized, it allows for easy experimentation with different granularities for the discretizations of the DNN output. The results are shown in table~\ref{tab:assumption_results}.

%For instance, when MaxCTE=4 (i.e., 5 intervals for \vcte) the size of the assumption is 13 states while $M_1$ has 261 states. When, MaxCTE=100 means that \vcte has 101 intervals while \he has 3 intervals, thus the DNN can be seen as having $101*3=303$ outputs. The generated assumption is still reasonable (301 states) and the assumption generation is still fast on my very old computer.

%The results are encouraging, showing that the assumption generation is effective even when the interface, corresponding to the size of the DNN output, is large.

%Current approaches for assumption generation \cite{ASE02} are effective when interfaces are small. 
%We note that our approach is general, and 
Our approach is general, and is not dependent on the granularity of  discretization used for the system states ($\vcte$ and $\vhe$); however, this granularity defines the size of the interface alphabet and can thus affect the scalability of the approach. 

%To evaluate, we %performed an experiment generated assumptions for TaxiNet under different sizes of the interface alphabet ($\Sigma=Est$). 
We experimented with generating assumptions $\weakest{Est}$ for the TaxiNet case study, under different alphabet sizes; i.e., different values of \fspsyn{MaxCTE} defining the granularity for $\vcte$; the granularity of $\vhe$ stays the same. Table~\ref{tab:assumption_results} shows the results; we used an implementation in LTSA on a MacBook Air.
%\corina{computer configuration }

We first note that the generated assumptions are much smaller than the corresponding $M_1$ components. For instance, for \fspsyn{MaxCTE=2},  $M_1$ has 99 states (and 155 transitions) while the assumption is much smaller (7 states); it appears for this problem, the assumption size is linear in the size of the interface alphabet, making them good candidates for efficient run-time monitoring. The results indicate that the assumption generation is effective even when the size of the interface alphabet---corresponding to the number of possible DNN output values---is large. For instance, when \fspsyn{MaxCTE=100}, it means that $\vcte$ has 101 intervals while $\he$ has 3 intervals, thus
the DNN can be seen as having $101*3=303$ possible discrete output values. The generated assumption has 301 states and the assumption generation is reasonably fast. 
%\corina{Similar trends observed for the other assumption}
The results indicate that our approach is promising in handling practical applications, even for DNNs (classifiers) with hundreds of possible output values.
%
%If the output dimension of the DNN is much larger the assumption generation may no longer work. 
%We can try to group together multiple predictions guided by the logic of the downsteam decision making (similar to what we have done for the TaxiNet example). 

In case assumption generation no longer scales, 
we can group multiple DNN output values into a single (abstract) value, guided by the logic of the downstream decision making components (similar to how we group together multiple continuous DNN output values into discrete values representing intervals in the TaxiNet example). 
Incremental techniques, that use learning and alphabet refinement \cite{GheorghiuGP07} can also help alleviate the problem and we plan to explore them in the future.

%\subsection
\paragraph{\textbf{Assumptions as Run-time Safety Monitors}}

%\corina{Analysis with PRISM}

%We explore the use of the automatically generated assumptions as run-time safety monitors that guarantee the satisfaction of the required safety properties by the autonomous system. 
The goal of this evaluation is to: (i) check that the TaxiNet system augmented with the safety monitor is guaranteed to be safe, (ii) quantify the permissiveness of the monitor, i.e., the probability of the the assumption being violated during system operation.

To this end, we devised an experiment that leverages probabilistic model checking using the $\prism$ tool~\cite{PRISM}. We built $\prism$ models for the TaxiNet $Controller$ and $Dynamics$ components that are equivalent to the corresponding LTSs encoded in the FSP language.
%closely match the logic encoded in FSP. For perception, we built a conservative abstraction that maps every actual value to every possible estimated value; its probabilistic transitions are empirically derived from running the DNN on a representative data set. 
We also had to encode the $Sensor$ and $Perception$ components since our goal is to study the behavior of the overall system.
For this purpose, we use our prior work~\cite{pasareanu2023closedloop} to build a conservative probabilistic abstraction of $M_2 = Sensor\parop Perception$ that maps every actual system state value to a probability distribution over estimated system state values; the probabilities associated with the transitions from actual to estimated values are empirically derived from running the DNN on a representative data set provided by the industrial partner (11,108 images). 

%The logic of the monitor closely implements the transitions of the assumption, with an additional abort (or fail-safe) state, that traps the output behaviours of the DNN that can potentially lead to safety violations in the overall system.

\begin{figure}[t]
\centering
  \centering
  \includegraphics[width=0.9\linewidth]{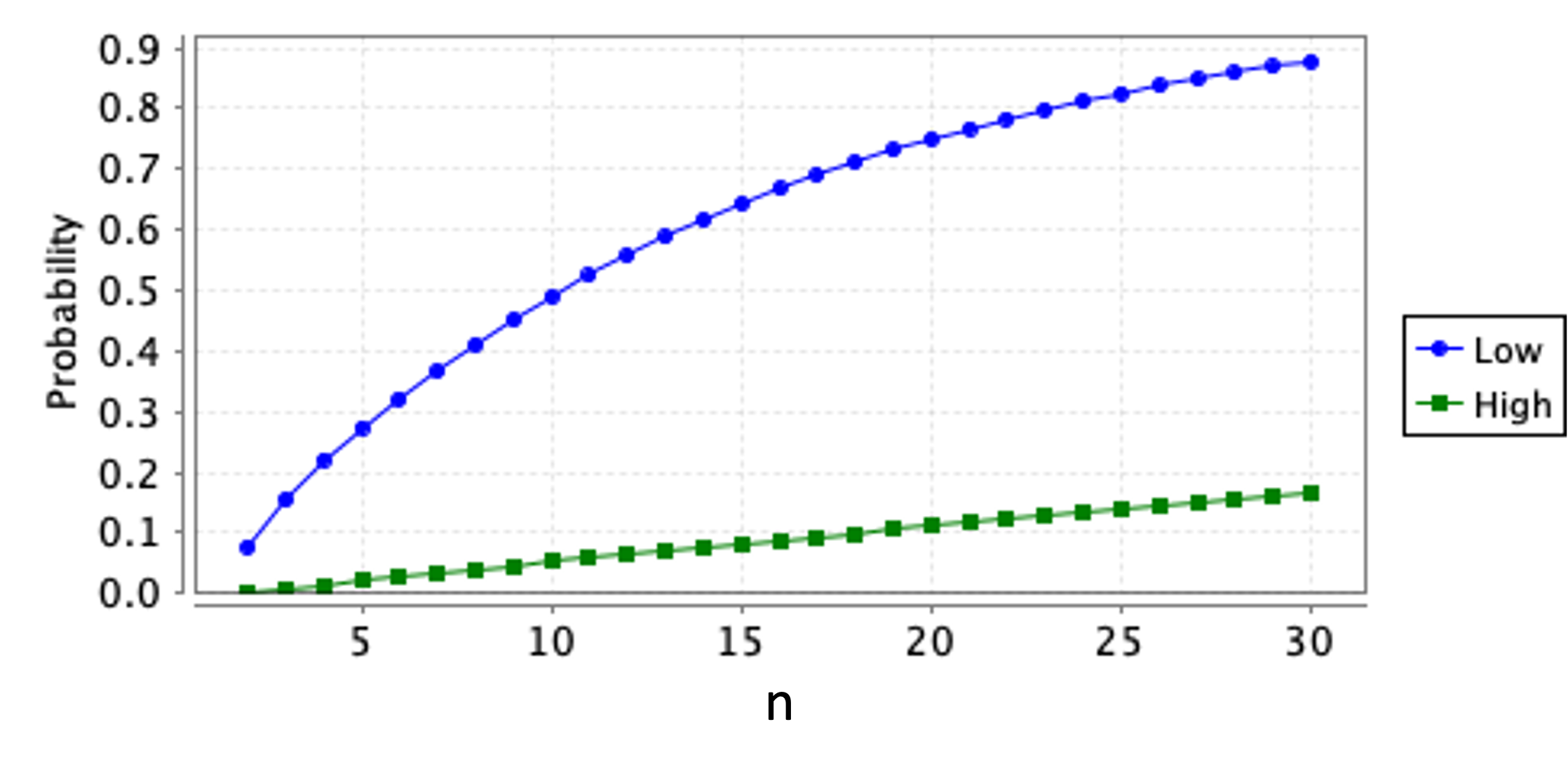}
  \captionof{figure}{Probability of the assumption being violated. $n$ indicates horizon length. Low and High correspond to lower and higher accuracy DNNs.}\vspace{-0.7cm}
  \label{fig:results}
\end{figure}

 \noindent{\bf PRISM results.} We first double-checked that the $\prism$ model of the TaxiNet system, augmented with the run-time monitor\footnote{We provide the code of the monitor in the appendix.}, does not violate the two safety properties, which $\prism$ confirmed, validating the correctness of our approach.  We also analyzed a PCTL \cite{PRISM} property, $P=? [ F (Q=-1)]$, that asks for the probability of the system reaching  \fspsyn{Q=-1}, which encodes the error state of the assumption.
 %thereby quantifying the permissiveness of the run-time monitor. 
 The results for this property are shown in Figure~\ref{fig:results} for two different versions of the DNN that vary in their accuracies. While the probability of violating the assumption as the horizon length increases tends towards 1 for both the DNNs, the rate of growth for the higher accuracy DNN is much slower. Developers can use the analysis to evaluate the permissiveness of the run-time monitor. 
 %the quality of the trained DNN from the perspective of abort frequency. %\corina{comment a bit}.

\section{Related Work}

There are several approaches for formally proving safety properties of autonomous systems with low-dimensional sensor readings ~\cite{ivanov2021compositional,ivanov2019verisig,ivanov2020verifying,ivanov2021verisig,dawson2022learning,dawson2022safe}; however, they are  intractable for systems that use rich sensors producing high-dimensional inputs such as images. 
%
%This is due to the difficulty of precisely modeling the camera~\cite{santa2022nnlander}, the external environmental conditions that are difficult to account for. Second, even if one manages to construct a mathematical model of the image generator (i.e., camera),  the combined complexity of it and the neural networks used for perception makes current closed-loop analysis algorithms prohibitively expensive. For instance, 
%
%NNLander-VeriF~\cite{santa2022nnlander} builds mathematical models of the camera and combines them with the neural network and the rest of the system, but can only handle small images and networks.
%black and white images with $16\times 16$ pixels and small networks.  
More closely related works aim to build models based on the analysis of the perception components. However, they either do not provide guarantees~\cite{katz2022verification} or do not scale to large networks~\cite{santa2022nnlander}.

The most closely related approach is the one in \cite{hsieh2022verifying}, which builds abstractions of the DNN components as guided by system-level safety properties. %The main difference from our work is that  their abstractions are not guaranteed to be conservative approximations of the perception components, and thus the 
The method does not provide strong system-level guarantees; instead it only provides a probabilistic result that measures empirically how {\em close} a real DNN is to the abstraction. Another difference is that 
%our assumptions are in terms of {\em sequences} of DNN outputs, and are thus potentially more expressive. Furthermore, 
%in our construction we do not rely on the training data, whereas 
the approach in ~\cite{hsieh2022verifying} uses the training data to help discover the right abstraction, whereas we do not rely on any data.
%\ravi{Relate the UIUC paper to local specs}

In recent work ~\cite{pasareanu2023closedloop}, we built a probabilistic abstraction of the camera and perception DNN for the probabilistic analysis of the same TaxiNet case study. 
%This enables scalable closed-loop analysis for systems with a DNN in the loop. 
The approach facilitates obtaining {\em probabilistic} guarantees with respect to the satisfaction of safety properties of the entire system. Another recent work leverages assume-guarantee contracts and probabilistic model checking to evaluate probabilistic properties of the end-to-end autonomy stack~\cite{incer2023pacti}. In contrast, we focus here on obtaining strong (non-probabilistic) safety guarantees.

The work in~\cite{P.Habeeb2023} aims to verify the safety of
the trajectories of a camera-based autonomous vehicle in a given 3D-scene. 
%The work use invariant regions over the input space grouped based on the same controller action. However, 
Their abstraction  captures only one environment condition (i.e., one scene) and one camera model, whereas our approach is not particular to any camera model and implicitly considers all the possible environment conditions.

%\divya{Added the paragraph below. Please check.}
%All these techniques are specific to a given DNN model. They either employ the network as is in the analysis, or rely on the behavior of the model on a specific set of data to build an abstraction. Our approach, on the other hand, is performed in the absence of a DNN, and it thus does not rely on knowledge about the DNN or access to the training data. 

In a previous white paper~\cite{pasareanu2018comp}, we advocated for a compositional framework with input-output DNN contracts obtained from a DNN-specific analysis. However, in that work, we left open the problem of how to precisely relate the contracts to the system level properties. We solve that problem here, where instead of input-output contracts for a DNN, we derive assumptions that are based solely on the outputs of the DNN. The assumptions are derived without a DNN-specific analysis.

%VerifAI~\cite{ghosh2021counterexample} is another framework that uses compositional techniques for the analysis of leraning-enabled systems, however it can only find counter-examples to system safety, so it can not provide guarantees.

Our work is also related to safe shielding for reinforcement learning \cite{alshiekh2017safe}. %The goal there is to monitor the actions taken by a controller and replace them with safe actions in case of violations of system-level safety properties. 
That work does not consider complex DNNs as part of the system and therefore does not discuss suitable techniques for them. Nevertheless, we note that our assumptions are monitoring the outputs of the DNN instead of the actions of the controller (as in shielding), and can thus be used to prevent errors earlier. They also act as local specifications for the DNN behaviour, enabling other activities such as testing or training.

\section{Conclusion and Future Work}

We presented a compositional approach based on automated assumption generation
%assume-guarantee reasoning 
for the verification of autonomous systems that use DNNs for perception. We demonstrated our approach on the TaxiNet case study. While our %compositional 
approach opens the door to analyzing autonomous systems with state-of-the-art DNNs, it can suffer from the well known scalability issues associated with model checking (due to state-explosion). We believe we can address this issue via judicious use of abstraction and compositional techniques. 
%For instance, the analysis of $Controller\parop Dynamics$ can itself be further decomposed into the separate analysis of the $Controller$ and $Dynamics$, while each component can be made amenable for verification using further abstraction techniques. 
Incremental, more scalable, techniques for assumption generation can also be explored, see. e.g.~\cite{tacas03}.

%Current approach for assumption generation is effective when interfaces are small. If the output dimension of the DNN is very large (e.g., 100 classes for Cifar100) the proposed method is likely not to work. In that case, we can try to group together multiple predictions guided by the logic of the downsteam decision making (similar to what we have done for the TaxiNet example). Incremental techniques, the use learning and alphabet refinement \cite{?} can also help alleviate the problem and we plan to explore them in the future.

We have presented our approach in the context of components %($Controller$ and $Dynamics$) 
modeled as LTSs. However, such components are often modeled as more complex hybrid automata. 
We believe that our proposed approach can be extended to reasoning about such systems, leveraging our previous work on assumption generation for hybrid automata \cite{BogomolovFGGPPS14}.
%, albeit that work did not consider learning-enabled components.

Another direction for future work is to investigate autonomous systems with multiple machine learning components (e.g., using both camera and LIDAR for sensing) and to develop techniques that decompose the generated global assumption into  assumptions for each such component. These assumptions can then be used to guide the development of the components and can also be deployed as component-specific run-time monitors.

%\corina{say something about addressing regression DNN models directly; also object detectors}

\newpage
%%
%% The acknowledgments section is defined using the "acks" environment
%% (and NOT an unnumbered section). This ensures the proper
%% identification of the section in the article metadata, and the
%% consistent spelling of the heading.
%\begin{acks}
%\end{acks}

%%
%% The next two lines define the bibliography style to be used, and
%% the bibliography file.
\bibliographystyle{IEEEtran}
\bibliography{bibfile}

\newpage
%% Appendix
%\appendices
\appendix
%\section{FSP Code for Pessimistic TaxiNet LTS}
%\label{sec:code_ltsa_pessimistic}

%\section{FSP Code for Optimistic TaxiNet LTS}
%\label{sec:code_ltsa_optimistic}

\subsection{$\prism$ Encoding for TaxiNet with Safety Monitor}
\label{sec:code_prism}

We show the $\prism$ code for $M_2$ and the safety monitor in Figure~\ref{fig:taxinet-prism}.
We use the output of step 4 in procedure \Fbuildassume (Section~\ref{sec:background_assumption})
as a safety monitor, i.e., the assumption LTS has both $err$ and $sink$ states, with a transition to $err$ state interpreted as the system aborting.

The $\prism$ encoding of the safety monitor closely follows the transitions of the assumption computed for $M_1$ over alphabet $\interalpha=Est$.
%, with an additional $abort$ (or fail-safe) state added to trap the output behaviours of the DNN that violate the assumption and potentially lead to safety violations of the overall system.

\begin{figure}%[!t]
\centering
\input{code/code_prism.tex}
\caption{TaxiNet $M_2$ and safety monitor in $\prism$. }
\label{fig:taxinet-prism}
\end{figure}
%\textbf{$\prism$ Encoding.} We show the $\prism$ code for $M_1$ and the safety monitor in Figure~\ref{fig:taxinet-prism}.
%The complete $\prism$ code for the TaxiNet system is shown in Appendix~\ref{sec:code_prism}. 
%
In the code, variable \fspsyn{pc} encodes a program counter. 
%The (imperfect) perception maps the
$M_2$ is encoded as mapping the actual system state (represented with variables $\vcte$ and $\vhe$) to different estimated states (represented with variables \fspsyn{cte\_est} and \fspsyn{he\_est}). The transition probabilities are empirically estimated based on profiling the DNN; for simplicity we update \fspsyn{cte\_est} and \fspsyn{he\_est} in sequence. The monitor maintains its state using variable \fspsyn{Q} (initially \fspsyn{0}); it transitions to its next state after \fspsyn{cte\_est} and \fspsyn{he\_est} have been updated;  the abort state (\fspsyn{Q=-1}) traps behaviours that are not allowed by the assumption; there are no outgoing transitions from such an abort state. 
%The controller and the dynamics only advance if {\tt pc=4} and {\tt pc=5}, respectively.

\end{document}